# A Unified Framework for Order-of-Magnitude Confidence Relations

**Didier Dubois**
IRIT
118 route de Narbonne
31062 Toulouse, France

**Hélène Fargier**
IRIT
118 route de Narbonne
31062 Toulouse, France

## Abstract

The aim of this work is to provide a unified framework for ordinal representations of uncertainty lying at the crossroads between possibility and probability theories. Such confidence relations between events are commonly found in nonmonotonic reasoning, inconsistency management, or qualitative decision theory. They start either from probability theory, making it more qualitative, or from possibility theory, making it more expressive. We show these two trends converge to a class of genuine probability relations, numerically representable, that cumulate features of probability and possibility theories. We provide characterization results for these useful tools that preserve the qualitative nature of possibility rankings, while enjoying the power of expressivity of additive representations.

## 1 Introduction

Comparative probability emerged in the thirties with the works of Ramsey, Koopman and De Finetti (see the anthology edited by Kyburg and Smokler, 1980) with a view to provide subjectivist foundations to probability theory. Under this view, the notion of belief, expressed in terms of probability, is comparative prior to being numerical, numerical probabilities being a matter of measurement of a basically ordinal notion. In the last twenty years, the modelling of relative belief has been often encountered in the knowledge representation and reasoning literature under the form of a nonprobabilistic relation on a set of propositions or events. This is clearly the case with the nonmonotonic reasoning literature especially the so-called preferential models and inference (Shoham 1988, Lehmann and Magidor 1992, Boutilier 1994, Friedman and Halpern 1996), and the related field of belief revision (see e.g. Rott 2001). In both fields, the confidence relation turned out to be closely related to the comparative possibility relation (see e.g. Lewis (1973) or Dubois (1986)). And indeed the axiomatic framework of Lehmann and Magidor as well as the axioms of belief revision enforce the existence of a possibility relation or a family thereof, ordering the set of propositions in terms of relative belief or epistemic entrenchment.

A particularly typical feature of these possibility orderings and their relatives is the following: they embed a idea of negligibility, first stressed by Lehmann (1996). If a proposition $A$ is more plausible than $B$ and also more plausible than $C$, then $A$ is more plausible than the disjunction of $B$ and $C$. It means that the plausibility of $B$ and that of $C$ are of a much lower order of magnitude than the one of $A$.

The strong conflict existing between probability theory and belief structures deriving from the knowledge representation and reasoning perspective has led to a rebuttal of nonmonotonic reasoning, based on the lottery paradox (Kyburg, 1988, Poole, 1991). However Paul Snow (1999) and others (Benferhat et al, 1999) showed that possibility and probability structures were not totally inconsistent with each other. They could lay bare a special kind of regular probability measures displaying the negligibility properties: linear big-stepped probabilities. On the other hand, as noticed in Dubois et al. (1998), Fargier and Sabbadin (2003), possibility relations can be usefully refined by lexicographic schemes and such refinements recover all properties of probability relations.

The aim of this paper is to provide a unified framework for ordinal representations of uncertainty lying at the crossroads between possibility and probability theories. We can then start either from probability theory, making it more qualitative, or from possibility theory, making it more expressive. We show these two trends converge to a class of genuine probability relations, numerically representable, called big-stepped probabilities, that cumulate features of probability and possibility theories. We thus build a bridge between ordinal belief structures, possibility and probability, by providing a framework that combines additivity and negligibility, thus organizing the existing concepts into a clear setting. This approach preserves the qualitative nature of possibility rankings, while enjoying the power of expressivity of additive representations.

Section 2 present a minimal ordinal setting for representing comparative confidence in events, that can serve for both probabilistic and possibilistic representations. Section 3 provides examples of confidence relations found in the literature, that bridge the gap between the additivity of probability theory and the negligibility effect at work in possibility theory. Section 4 presents axiom systems that lead to these relations. Proofs are omitted for the sake of brevity.

## 2 An Ordinal Setting for Confidence Relations

### 2.1 Confidence Functions and Relations

Confidence in events can be modelled by a monotone set-function on $S$, that is, a mapping $\mu$ defined from $2^S$ to $[0,1]$ such that:
- $\mu(\emptyset) = 0$, $\mu(S) = 1$,
- $\forall A, B \subseteq S, A \subseteq B \implies \mu(A) \leq \mu(B)$

We will call these functions *confidence functions*[1].

Important subclasses of confidence functions are probability and possibility/necessity measures. These measures are simple because the set-functions comparing the relative likelihood of events are completely determined by the knowledge of the degrees of confidence in states (either a probability distribution $p$ or a possibility distribution $\pi$ on $S$):
- $P(A) = \sum_{s \in A} p(s)$,
- $\Pi(A) = \max_{s \in A} \pi(s)$, $N(A) = 1 - \max_{s \in \bar{A}} \pi(s)$

A more general approach is to use a relation that compares propositions in terms of the relative confidence an agent possesses about them. The basic properties of confidence measures extend to such partial relations we call *confidence relations* (Dubois et al. 2003) :

**Definition 1** *Let $\succeq$ be a relation on $2^S$, $\sim$ its symmetric part, $\succ$ its asymmetric part. $\succeq$ is a confidence relation iff it is:*

- *Reflexive:* $\forall A, A \succeq A$
- *Non trivial:* $S \succ \emptyset$
- *Consistent:* $\forall A, S \succeq A$ and $A \succeq \emptyset$
- *Quasi-transitive:* $A \succ B$ and $B \succ C \implies A \succ C$
- *Monotonic ("orderly" after Halpern, 1997):*
  $\forall A, B, C \subseteq S$:
  $A \succeq B \Rightarrow A \cup C \succeq B$ and $A \succeq B \cup C \Rightarrow A \succeq B$

*If $\succeq$ is moreover a weak order (i.e. is complete and transitive), it is called a* confidence weak order.

The monotony of $\succeq$ implies the one of $\succ$; importantly, it also implies the ordinal counterpart to the set-function monotonicity:

**Proposition 1** *If $\succeq$ on $2^S$ is a monotonic confidence relation then $\forall A, B, C \subseteq S$:*
- $A \succ B \Rightarrow A \cup C \succ B$, $A \succ B \cup C \Rightarrow A \succ B$
- $A \subseteq B \Rightarrow B \succeq A$

Any capacity obviously defines a confidence weak order and conversely: $A \succeq_\mu B \iff \mu(A) \geq \mu(B)$. But not all confidence relations can be represented by a confidence function. It may be so because the relation is not complete (some events may be not comparable), as for instance generalized qualitative probabilities (Lehmann 1996) or that the indifference relation induced by relation $\succeq$ is not transitive. It accounts for some indistinguishability threshold in the perception of relative confidence by an agent. Two events may be declared as equally likely because the perception of the agent is coarse. Then, confidence levels are subject to Poincaré paradox: one may have $A \approx B, B \approx C$, but $A \succ C$. Moreover, some of the confidence relations that naturally pop up in our framework are naturally quasi-transitive.

### 2.2 CP-Relations and OM-Relations

Confidence functions and relations distinguish two important but generally incompatible properties: preadditivity on the one hand, that ensures that events are compared "ceteris paribus", and negligibility on the other hand, that is the peculiarity of many qualitative representation frameworks of AI (e.g. possibility relations, kappa functions, Lehmann's generalized probabilities, infinitesimal probabilities, and the like).

**Definition 2 (Preadditivity)** *A relation $\succeq$ on $2^S$ is preadditive iff it satisfies the following axiom:*
**ADD**: $\forall A, B, C \subseteq S$ such that $A \cap (B \cup C) = \emptyset$:
$B \succeq C \Leftrightarrow A \cup B \succeq A \cup C$.

The rationale behind this property, first requested by De Finetti, is clear: the part common to sets $A \cup C$ and $A \cup B$ should be immaterial for their comparison in terms of relative confidence. It was introduced as

---
[1] They have been given various names such as capacities (Choquet), fuzzy measures (Sugeno) or plausibility measures (Halpern) in various domains. The latter denomination suits confidence modelling but could be confused with Shafer's plausibilities that have a more precise meaning, as a weak counterpart to the notion of belief – they are particular monotone set-functions. The term "confidence function" intends to avoid ambiguity and remain open

the natural counterpart to the additivity property of probability functions (see e.g. Fishburn, 1986). In the same way, we can define CP-relations as a generalization of comparative probabilities:

**Definition 3** *A comparative probability relation is a transitive, complete and preadditive monotonic confidence relation.*
*A ceteris paribus confidence relation (or CP-relation for short) is a preadditive confidence relation.*

For instance, the partial ordering of events generated by a family of probabilities $\mathcal{F}$: $A \succeq B$ iff $P(A) \geq P(B), \forall P \in \mathcal{F}$ is a (non-complete) CP-relation.

Notice that comparative probabilities are more general than the relations induced by usual probability measures, as shown by Kraft's et al. (1959) counterexample. In addition to probability measures, comparative probabilities encompass relations not representable by a classical probability, only by means of special classes of belief functions. This is why we call the property in Definition 2 preadditivity.

The second noticeable property is Negligibility, that usually comes along with Closeness, thus comparing events on the basis of their orders of magnitude:

**Definition 4** *A monotonic relation $\succeq$ on $2^S$ is an order of magnitude confidence relation (OM-relation) iff strict part satisfies the negligibility axiom:*

*Axiom* **NEG**: $\forall A, B, C \subseteq S$, *pairwise disjoint sets,*
$A \succ B$ *and* $A \succ C \implies A \succ B \cup C$

*and its symmetric part is a closeness relation i.e., it satisfies*

*Axiom* **CLO**: $\forall A, B, C \subseteq S$
$A \sim B$ *and* $(A \succ C$ *or* $A \sim C) \implies A \sim B \cup C$.

An event is close to another iff their plausibilities have the same order of magnitude: a set is obviously close to itself, and to any union of sets of the same order of magnitude. Contrary to closeness, negligibility deals with *disjoint* sets. Axiom NEG states that, if $B$ and $C$ are negligible w.r.t. $A$, then so is also $B \cup C$. This feature looks quite counterintuitive in probabilistic terms, but is at the foundation of many uncertainty frameworks proposed by AI e.g. kappa or possibility functions and similar models. It makes sense in the scope of nonmonotonic reasoning and belief revision where the notion of belief is often interpreted as *tentatively accepted belief until new information is available*. In belief revision theory (e.g. Rott, 2001) belief sets are deductively closed, and thus the conjunction of two beliefs is a belief. This requirement makes sense for accepted beliefs, and is used in the preferential inference approach to nonmonotonic reasoning (Kraus et al, 1990). It is precisely this requirement that enforces the idea of negligibility in the underlying model of relative belief (Dubois et al 2004). The closure under conjunction leads to the following axiom (Dubois and Prade, 1995, Friedman and Halpern 1996):

Axiom **CCS**: $\forall A, B, C$ pairwise disjoint subsets of $S$,
$A \cup C \succ B$ and $A \cup B \succ C \implies A \succ B \cup C$

Axiom CCS also called "union property" by Halpern (1997). We could use it to enforce negligibility (Dubois et al. 2004). But NEG is actually less demanding than CCS and already expresses negligibility. Halpern (1997) also strengthens the CCS axiom thus into "Qualitativeness" axiom:

Axiom **QUAL**: $\forall A, B, C \subseteq S$,
$A \cup C \succ B$ and $A \cup B \succ C \implies A \succ B \cup C$

QUAL is a very strong condition, since it not only implies CO, NEG and its extension to non disjoint sets, but also the transitivity of $\succeq$. In Definition 4, we propose to build order of magnitude relations on weaker requirements.

Typical examples of negligibility relations are comparative possibilities (Lewis, 1973) and comparative necessities, that use a totally ordered scale of magnitude:

**Definition 5** *A comparative possibility (resp. necessity) is a confidence weak order $\succeq_\Pi$ such that:*
$\forall A, B, C \subseteq S, (B \succeq_\Pi C \implies A \cup B \succeq_\Pi A \cup C)$
*(resp. $A \cup B \succeq_N A \cup C \implies B \succeq_N C$)*

Possibility orderings have been studied by Grove (1988) in the scope of belief revision. Interestingly, it has been shown (Dubois, 1986) that comparative possibility relations are equivalently represented by the corresponding set-functions and thus derive both from a weak order $\succeq_\pi$ on states only. Intuitively $s \succeq_\pi s'$ means that state $s$ is at least as plausible, normal, expected as state $s'$. The main idea behind this modelling is that the state of the world is always believed to be as normal as possible, neglecting less normal states. It presupposes that when assessing the comparative plausibility of events, an agent focuses on the most normal situations and neglects other less plausible ones.

Comparative possibilities are OM-relations: possibility levels purely express orders of magnitude of plausibility. But comparative necessities are not. The closeness property is indeed not satisfied (since e.g. $N(A) = \emptyset$ and $N(B) = \emptyset$ does not imply $N(A \cup B) = \emptyset$). Necessity relations express comparative levels of certainty or belief (a notion stronger than plausibility).

Our first result is that negligibility is not compatible with preadditivity except in very particular cases:

**Theorem 1** *If $\succeq$ is both a comparative probability and satisfies CLO, then there exists a permutation of the elements of S such that:*
$\{s_1\} \succ \{s_2\} \succ \ldots \{s_k\} \succ \emptyset$ and $\{s_{k+1}\} \approx \{s_{k+2}\} \approx \cdots \approx \{s_n\} \approx \emptyset$

*If $\succeq$ is both a comparative probability and satisfies NEG, then there exists a permutation of the elements of S such that:*
$\{s_1\} \succ \{s_2\} \succ \ldots \{s_k\} \succ \{s_{k+1}\} \succeq \{s_{k+2}\} \succeq \{s_{k+3}\} \succ \emptyset$ and $\{s_{k+4}\} \approx \cdots \approx \{s_n\} \approx \emptyset$

Stronger variants of this theorem appear in Dubois et al. (2003, 2004). So, requiring preadditivity within possibility theory, or equivalently, requiring that a comparative probability reflects order of magnitude reasoning leaves little room: either possibility relations based on linear distributions, or, which is equivalent, linear big-stepped probabilities (i.e. probability functions such that there exists a permutation of the states where $\forall i, P(\{s_i\}) > \sum_{j>i} P(\{s_j\})$. In particular it means that two states of the world cannot be equally plausible. Families of such linear orders are at work in the nonmonotonic System P of Kraus et al., 1990 (see Benferhat et al., 1999).

One may wonder whether there exist confidence relations that cumulate the advantages of preadditivity and the qualitativeness of order of magnitude reasoning. The question is not so paradoxical as it may seem to be: Section 3 shows preadditive relations escaping Theorem 1 by relaxing one of its conditions, namely either the transitivity of the confidence relation or the scope of the negligibility and closeness axioms.

## 3 Examples of Preadditive Relations Involving Negligibility

In this section, we start from possibility relations and try to extend their discrimination power by injecting preadditivity. Two events $A$ and $B$ may fail to be discriminated using the dual pair $(\succeq_N, \succeq_\Pi)$ as soon as $A \approx_\Pi B$ and $A \approx_N B$, (this is the so-called "drowning effect"). $A \approx_\Pi B$ may be due to a high plausibility of $A \cap B$. However following the preadditivity property $A \cap B$ should not affect the comparison between $A$ and $B$, this set being common to both. The same reasoning applies to $\bar{A} \cap \bar{B}$ with respect to the equality $A \approx_N B$. In the spirit of probability theory, only $\bar{B} \cap A$ and $B \cap \bar{A}$ should matter in telling $A$ from B.

### 3.1 Discrimax Possibilistic Likelihood

Hopefully, there exits a preadditive refinement of both $\succeq_N$ and $\succeq_\Pi$, where events are compared only with respect to their disjoint part:

**Definition 6 (Dubois et al. 1998)** *Let $\succeq_\Pi$ be a comparative possibility relation on $2^S$. The discrimax likelihood relation (denoted $\succeq_{\Pi L}$) is defined as follows:*

$\forall A, B \in 2^S, A \succeq_{\Pi L} B \iff A \cap \bar{B} \succeq_\Pi \bar{A} \cap B$

This kind of construct can be found in many works (surveyed in (Halpern, 1997)) mainly for the purpose of ordering consistent subsets of formulas from a prioritized knowledge base, in the scope of nonmonotonic reasoning and inconsistency management. It also appears as the *only* rational Savagean model of uncertainty under the assumption of a purely ordinal evaluation of decisions (Dubois et al, 2003)

$\succeq_{\Pi L}$ is obviously a CP-relation. The comparison of two events is purely ordinal and is very simply obtained from a possibility distribution. Indeed:

$A \succeq_{\Pi L} B \iff max_{s \in A \cap \bar{B}} \pi(s) \geq max_{s \in \bar{A} \cap B} \pi(s)$

The latter formulation motivates the name "discrimax", short for "maximal plausibility on discriminating sets". $\succeq_{\Pi L}$ is akin to an OM-relation in the following sense:

**Proposition 2** *The restriction of $\succeq_{\Pi L}$ to disjoint events is an OM-relation, i.e. for $A, B, C$ pairwise disjoint sets:*
$A \succ_{\Pi L} B$ and $A \succ_{\Pi L} C \Rightarrow A \succ_{\Pi L} B \cup C$
$A \sim_{\Pi L} B$ and $(A \sim_{\Pi L} C$ or $A \succ_{\Pi L} C) \Rightarrow A \sim_{\Pi L} B \cup C$

Like comparative probabilities, discrimax likelihood is preadditive and thus escapes the main drawback of possibility theory (the drowning effect). But it stays fully compatible with the possibilistic framework since it refines both $\succeq_\Pi$ and $\succeq_N$: when $A$ is strictly more possible (or more necessary) than $B$, it is also judged more likely than $B$. Formally :

**Proposition 3** *(Dubois et al., 1998)*
$A \succ_\Pi B \implies A \succ_{\Pi L} B; A \succ_N B \implies A \succ_{\Pi L} B$
$A \cap B = \emptyset \implies (A \succeq_{\Pi L} B \iff A \succeq_\Pi B)$

But it may happen that, whereas $A$ and $B$ are equally possible (resp. necessary) since relying on the same very possible states (resp. excluding the same impossible states), the former is judged more likely than the second, because, apart from these common elements, the possibility of $B \cap \bar{A}$ is far lower than the one of $A$.

It should also be noticed that $\succeq_{\Pi L}$ is quasi-transitive: its strict part is transitive, but it may happen that $A \sim_{\Pi L} B, B \sim_{\Pi L} C$ whereas $A \succ_{\Pi L} C$. This is due to the fact that $A \cap B = \emptyset$ and $B \cap C = \emptyset$ do not imply $A \cap C = \emptyset$.

## 3.2 Big-Stepped Probabilities

This second model (Dubois et al., 1998; Fargier and Sabbadin, 2003) is particularly appealing, in the sense that it proposes a form of comparative probability compatible with possibility theory:

**Definition 7** *A probability measure $P$ is said to be big-stepped iff:*

$$\forall s \in S, P(\{s\}) > P(\{s' \text{ s.t. } P(\{s'\}) < P(\{s\})\})$$

*i.e., for any $s$, $p(s) > \sum_{s' \text{ s.t. } p(s') < p(s)} p(s')$.*

It means that grouping together states of probability lower than the probability of a prescribed state does not form an event more probable than this state. According to this definition, any probability measure defined by a uniform distribution on $S$ is formally big-stepped since the condition vacuously holds. Whether this probability is genuinely big-stepped is a matter of discussion, but it is a limit case thereof. This definition also generalizes the linear big-stepped probabilities of (Snow 1999, Benferhat et al. 1999) which are recovered when one moreover assumes that non null states are never equiprobable. The purely ordinal definition goes as follows:

**Definition 8** *A comparative probability $\succeq_P$ is big-stepped iff: $\forall s \in S, \{s\} \succ_P \{s' \text{ s.t. } \{s\} \succ_P \{s'\}\}$.*

A big-stepped comparative probability relation is always representable by a numerical (big-stepped) probability measure. Orderings of events induced by big-stepped probabilities are encountered in the AI literature. First, they have much in common with Spohn's $\kappa - functions$ (1988): disbelief degrees provided by a $\kappa - function$ can indeed be interpreted as the order of magnitude of a non-standard probability (Pearl, 1993; Giang and Shenoy 1999) — the latter non-standard probability is actually a big-stepped probability.

Big-stepped probabilities are fully compatible with possibility theory, in the sense that they refine discrimax possibilistic likelihood relations, hence possibility/necessity measures as well:

**Proposition 4** *Let $\pi$ be a possibility distribution and $p$ a probability distribution such that $\pi(s) \geq \pi(s') \iff p(s) \geq p(s')$. If $P$ is a big-stepped probability then: $A \succ_{\Pi L} B \implies A \succ_P B$.*

Indeed each cluster of equally probable states corresponds to a class of equipossible states, that is, equipossibility leads to equiprobability. When the possibility distribution on states is linear, the relations induced by big-stepped probabilities coincide with discrimax possibilistic likelihood relations. In the case of a uniform possibility distribution, the big-stepped probability is a regular uniform probability and the strict part of the discrimax possibilistic likelihood relation coincides with proper set-inclusion. Conversely, it is very easy to build the possibilistic structure that underlies a big-stepped probability:

**Proposition 5** *Let $P$ be a big-stepped probability. Then, there exists a possibility distribution $\pi$, ordinally equivalent to $p$ and such that $A \succ_\Pi B \implies A \succ_P B$.*

So, when an event is more possible (or more certain) than another, it is also more probable. But there are some equi-possible (resp. equi-necessary) events that are distinguished by the big-stepped probability, precisely those that suffer from the drowning effect: when the most plausible states realizing the two events are equally likely, they cancel each other in a pair-wise manner, and the comparison boils down to the comparison of the remaining states. Finally, if $A$ has more most plausible states than $B$ then $A$ is considered more likely. When the events have the same number of states of highest plausibility, the comparison then focuses on the second plausibility level, etc.

This mode of comparison is closely related to the lexicographic comparison of vectors called leximax:

**Definition 9 (Leximax ordering)** *Let $(L, \geq)$ be an ordered scale, $\vec{U}, \vec{V}$ two vectors of a space $L^N$, and $\sigma$ a permutation that ranks a vector by decreasing order ($\forall \vec{w} \in L^N, w_{\sigma(1)} \geq \cdots \geq w_{\sigma(N)}$: $\vec{U} \succeq_{leximax} \vec{V} \iff \exists k \in [1, N], u_{\sigma(i)} = v_{\sigma(i)}$ for $i < k$ and $u_{\sigma(k)} > v_{\sigma(k)}$.*

Now, let $pl : S \to [0, 1]$ be a distribution – e.g. a probability $p$, or a possibility distribution – encoding any confidence weak order on $S$, and call $\vec{A_{pl}}$ the confidence vector of $A$ ($a_i = pl(s_i)$ if $s_i \in A$, $a_i = 0$ otherwise).

**Theorem 2** *If $P$ is a big-stepped probability, then $\forall A, B \in 2^S, P(A) \geq P(B) \iff \vec{A_p} \succeq_{leximax} \vec{B_p}$ (for $pl = p$). Conversely, when $S$ is finite, $\forall pl$, there exists a big-stepped probability such that $\forall A, B \in 2^S, \vec{A_{pl}} \succeq_{leximax} \vec{B_{pl}} \iff P(A) \geq P(B)$.*

Since based on probabilities, big-stepped probabilities are obviously additive. But the numerical values are immaterial for characterizing the corresponding confidence relation. Only the confidence ordering of states matters, which confirms the ordinal nature of such comparative probabilities. Big-stepped probabilities satisfy the negligibility axiom only partially. Indeed, $A$ can be more probable than $B$ for two different reasons: either $B$ is negligible in front of $A$, or they are close to each other, i.e. their most probable states share the same probability, and then the difference is made by counting the number of states in the levels.

Only the restriction of the relation to events that are not close to each other satisfies NEG. But axiom CLO is not satisfied for big-stepped probabilities.

Moreover, big-stepped probabilities form a special class of *lexicographic probabilities* in the sense of (Blume et al., 1991; Lehmann 1998). Let $S_1, S_2, \ldots, S_n$ be a well-ordered partition of $S$. A lexicographic probability is a probability function whose restriction to events in the subalgebra generated by $S_1, S_2, \ldots, S_n$ is a linear big-stepped probability. It generates a possibility ordering on subsets made of union of $S_i$'s. But the subsets of each $S_i$ are ordered by a regular probability. In other terms, a lexicographic probability ordering is a probability function such that $\forall s \in S_i, \forall A \subseteq S_{i+1} \cup \cdots \cup S_n, p(s) > P(A)$. Notice that the restriction that all the states within a single cluster $S_i$ are equiprobable is dropped here and that big-stepped probabilities are recovered when the latter assumption is added. The idea of assigning probabilities to equipossible states, was also suggested by Boutilier (1994). Lexicographic probabilities obey the axioms of so-called generalized qualitative probabilities of Lehmann(1996), that are not supposed to be complete, and display negligibility effects.

Any discrimax likelihood relation induced by a possibility ordering of states $\succeq_\pi$ can also be generated by a family of lexicographic probability functions $\mathcal{F}$ sharing the same well-ordered partition induced by equipossible states. It is then easy to see that $A \succ_{\Pi L} B$ iff $P(A) > P(B), \forall P \in \mathcal{F}$.

## 4 Characterizing Confidence Relations that Exhibit Negligibility Effects

The core of the characterization of the relations of Section 3 is based on the assumption that, in a qualitative framework, the relation between states generates the whole relation among events – basically the one based on the order of magnitude of the states.

### 4.1 OM-relation induced by a quasi-transitive ordering of states

Halpern (1997) studied in detail the process of lifting a relation from elements of a set to its subsets, assuming a partial (pre)ordering $\succeq$ of states. Let us slightly enlarge these original assumptions: firstly, we start from a relation on $S \cup \{\emptyset\}$, so as to allow the representation of impossible states, and secondly we do not assume the transitivity of $\succeq$ but only its quasi-transitivity:

**Definition 10** *A basic confidence relation (basic relation, for short) is a reflexive and quasi-transitive relation $\succeq$ on $S \cup \{\emptyset\}$ that is:*

*consistent:* $\forall s, s \succeq \emptyset$
*non trivial :* $\exists s, \emptyset \succeq s$ *does not hold.*

These assumptions come up naturally in the setting of confidence relations. Indeed:

**Proposition 6** *Any confidence relation $\succeq$ induces a basic confidence relation $\succeq$: $s \succeq s' \Leftrightarrow \{s\} \succeq \{s'\}$.*

The basic relation corresponds to Halpern's ordering when it is supposed that $\forall s, s \succ \emptyset$ and, more importantly, that $\succeq$ is fully transitive (here, we only assume the transitivity of its strict part). Following Halpern, we define the strict confidence relation by:

$A \triangleright B \iff \forall s' \in B, \exists s \in A, s \succeq s'$ but not $s' \succeq s$.

But, contrary to (Halpern 1997) we do not use the same principle to build the symmetric part: since the full transitivity of $\succeq$ is not assumed here, it would lead to counterintuitive results. Here, we simply define the symmetric part by completing $\triangleright$ :

$A \doteq B \iff not(A \triangleright B)$ and $not(B \triangleright A))$

$A \trianglerighteq B \iff A \triangleright B$ or $A \doteq B$

Then the confidence relation $\trianglerighteq$ on events is said to be *simply generated* by the basic relation $\succeq$.

Notice that $A \trianglerighteq B$ is complete by definition. When $\succeq$ is also complete, the previous lifting principle implies:
$A \triangleright B \iff \forall s' \in B, \exists s \in A, s \succ s'$
$A \doteq B \iff \exists s_a \in A, s_b \in B, \forall s \in A, \forall s' \in B,$
$\quad s_a \succeq s'$ and $s_b \succeq s$

**Definition 11** *A relation $\trianglerighteq$ on $2^S$ is said to be consistent with a basic confidence relation $\succeq$ on $S$ iff:*
$\forall s,'s \in S, \{s\} \trianglerighteq \{s'\} \iff s \succeq s'; \{s\} \triangleright \emptyset \iff s \succ \emptyset$.

When $\succeq$ is complete (and quasi-transitive), as it is the case in each of the formalisms we aim at characterizing, it can be shown that there is only one OM-relation consistent with it, and that it is precisely the one described the previous lifting principle.

**Theorem 3** *Let $\succeq$ be a complete basic relation. The following propositions are equivalent:*
- $\trianglerighteq$ *is an OM-relation consistent with $\succeq$ ;*
- $\trianglerighteq$ *is simply generated by $\succeq$.*

So, $A \triangleright B$ when the order of magnitude of each state of $B$ is lower than the order of magnitude of some state in $A$ and $A \doteq B$ when the most plausible elements of $A$ and $B$ share the same order of magnitude.

The case when the OM-relation is a weak order, is obviously a characterization of comparative possibility.

**Theorem 4** *Let $\succeq$ be a monotonic confidence relation. The following properties are equivalent:*

- $\exists \pi$ such that $\forall A, B \subseteq S, A \succeq B \iff A \succeq_\Pi B$;
- $\succeq$ is a complete and transitive OM-relation.

Finally, we also conjecture that, when $\succcurlyeq$ is not complete, $\rhd$ is the least refined in the class of complete OM-relations in agreement $\succcurlyeq$. In this case, $A \rhd B$ iff it is explicitly stated by $\succcurlyeq$ (when $A$ is a singleton) or when $\forall s' \in B, \exists s \in A, s \succ s'$ or $s$ and $s'$ incomparable ($not(s \succcurlyeq s')$ and $not(s' \succcurlyeq s')$).

### 4.2 Axiomatics for Discrimax Possibility and Big-Stepped Probabilities

Confidence relations of Section 3 can be constructed from the basic confidence relation formed by their restriction to singletons – in these cases, this restriction is a weak order $\succcurlyeq$. But the relations under concern do not display the behavior of an OM-relation for all pairs of events. Nevertheless, the OM-relation $\rhd$ simply generated by $\succcurlyeq$ is at least embedded in the confidence relation in the sense of the following property:

**Compatibility of $\succeq$ with Order of Magnitude (COM)** $\forall A, B \subseteq S, A \rhd B \implies A \succ B$
where $\rhd$ is the OM-relation that consistent with $\succcurlyeq$.

COM ensures the minimal requirement stating that, if the order of magnitude of the confidence in $A$ is strictly higher than the one in $B$ (i.e. if $B$ is negligible w.r.t. $A$), then $A$ must be more likely than $B$.

An interesting case is when $\succeq$ is preadditive. The conjunction of COM and ADD induces ceteris paribus confidence relations with embedded negligibility, as appears in the discrimax likelihood model. In this case the restriction of $\succeq$ to disjoint subsets is an OM-relation, namely: :

**Ceteris Paribus Order of Magnitude (CPOM)**
$\forall A, B \subseteq S$ such that $A \cap B = \emptyset$, $A \rhd B \iff A \succeq B$

It is obvious that CPOM $\implies$ COM. If the preadditivity axiom is verified, the entire relation can be obtained from its restriction to disjoint events. Unfortunately, Theorem 1 shows that for OM-relations, NEG is not compatible with preadditivity when $\sim$ is transitive. Two different directions can be followed to obviate this conflict:

- Requiring the equivalence between $\succeq$ and $\rhd$ on disjoints sets only (CPOM). In this case, the transitivity of $\sim$ is lost.

- Requiring the transitivity of $\sim$ (i.e. a comparative probability). In this case, it is no more possible to ensure that $\sim$ is a closeness relation, even on disjoint sets. So, CPOM is relaxed into COM.

The following results show that, when the relation on singletons is a weak order, the first option is characteristic of the discrimax likelihood relation and, for the transitive case, big-stepped probabilities can be characterized by COM.

**Theorem 5** *Let $\succeq$ be a confidence relation. The following properties are equivalent:*
- *$\exists \pi$ such that $\forall A, B, A \succeq B \iff A \succeq_{\Pi L} B$*
- *$\succeq$ is preadditive, complete, satisfies CPOM, and its restriction to singletons is a weak order.*

**Theorem 6** *Let $\succeq$ be a monotonic confidence relation. The following properties are equivalent:*
- *There exists a big-stepped probability $P$ such that $\forall A, B, A \succeq B \iff P(A) \geq P(B)$*
- *$\succeq$ is a preadditive weak order and satisfies COM.*

Interestingly, the latter result shows that, while comparative probability relations are more general than (hence not representable by) numerical probabilities, it is when comparative probabilities get close to possibility relations, exhibiting negligibility effects, that they always have a quantitative representation (as big-stepped probabilities).

### 4.3 Lexicographic Probabilities

Lehmann (1996) defines generalized comparative probabilities that exhibit negligibility effects but escape the previous framework because they do not obey COM. Within this generalization of comparative probabilities, the negligibility relation is not built from a basic relation on states, like in possibility theory or Halpern's approach. Let us focus on the important class of probabilities suggested by Blume et al. (1991), called lexicographic probabilities, encountered in Section 3.2.

Let $\succcurlyeq$ be a lexicographic probability relation underlying a well-ordered partition $S_1, S_2, \ldots, S_n$ of $S$. Consider the restriction $\succcurlyeq_\mathcal{P}$ of $\succcurlyeq$ on $\mathcal{P} = \{S_1, S_2, \ldots, S_n\}$ as a (generalized) basic relation. An OM-relation $\rhd_\mathcal{P}$ can be simply generated from $\succcurlyeq_\mathcal{P}$ on a sub-algebra $\mathcal{B}$ of $2^S$ containing subsets of $S$ formed of unions of $S_i's$:

$A \rhd_\mathcal{P} B \Leftrightarrow \forall S_i \subseteq B, \exists S_j \subseteq A \; S_j \succ S_i.$

$\rhd_\mathcal{P}$ is then extended to all events $2^S$, as follows:

$A \rhd_\mathcal{P} B \iff A^* \rhd_\mathcal{P} B^*,$

where $A^* = \cup\{S_i, S_i \cap A \neq \emptyset\}$ is the upper approximation of $A$ in $\mathcal{B}$ in the sense of rough set theory. The integer $i = min\{j : S_j \cap A \neq \emptyset\}$ is the rank of $A$: the lower the rank of $A$ the higher the order of magnitude of its plausibility. The compatibility of a confidence relation $\succeq$ with the OM-relation $\rhd_\mathcal{P}$ then writes:

$COM_\mathcal{P} \; \forall A, B \subseteq S, A \rhd_\mathcal{P} B \implies A \succ B$

Lexicographic probabilities are obviously comparative probabilities that satisfy $COM_\mathcal{P}$. For regular comparative probabilities (exhibiting no negligibility effect), the subalgebra $\mathcal{B}$ reduces to $\{\emptyset, S\}$, and $A \rhd_\mathcal{P} B \iff A = S$ and $B = \emptyset$.

Conversely, it can be shown that, given a partition $\mathcal{P}$ of $S$, any comparative probability $\succeq$ satisfying $COM_\mathcal{P}$ is a lexicographic one. The lexicographic probability order can be reconstructed as follows: $\succcurlyeq_\mathcal{P}$ is the subpart of $\succeq$ that compares elements of $\mathcal{P}$ and each regular comparative probability $\succeq_i$ on $2^{S_i}$ is the restriction of $\succeq$ to subsets of $S_i$. It can then be shown that $A \succeq B$ if either $A^* \rhd_\mathcal{P} B^*$, or $A^* =_\mathcal{P} B^*$ (i.e. $A$ and $B$ have the same rank $i$), and then $A \cap S_i \succ_i B \cap S_i$. A question for further research is whether all complete generalized probabilities in the sense of Lehmann satisfy $COM_\mathcal{P}$. Also, relationships between OM- relations, infinitesimal probabilities and conditioning over sets of measure 0 look natural to investigate (see Halpern, 2001).

## 5 Conclusion

The aim of our work was to clearly lay bare connections between probability and possibility theories, by means of axiomatic characterizations and representation results. Unsurprisingly, lexicographic orderings lie at the crossroads. Casting preadditivity and negligibility inside the axiomatic decision theory framework, nonstandard utilities or big-stepped utilities are also obtained on top of probabilities. It leads either to decision criteria which are generalized Condorcet voting rules (Dubois et al., 2003), or to additive refinements of possibilistic criteria representable by (big-stepped) expected utility (Fargier and Sabbadin, 2003).